# Evaluator for Emotionally Consistent Chatbots

**Chenxiao Liu, Guanzhi Deng, Tao Ji, Difei Tang, Silai Zheng**

Viterbi School of Engineering, University of Southern California

## Abstract

One challenge for evaluating current sequence- or dialogue-level chatbots, such as Empathetic Open-domain Conversation Models, is to determine whether the chatbot performs in an emotionally consistent way. The most recent work only evaluates on the aspects of context coherence, language fluency, response diversity, or logical self-consistency between dialogues. This work proposes training an evaluator[1] to determine the emotional consistency of chatbots.[2]

## 1 Introduction

### 1.1 Research Background

In the current research, chatbots can recognize and acknowledge a speaker's feelings in a conversation. However, we are still facing a big problem about how to ensure that chatbots are emotionally consistent. A practical and qualified chatbot can not only respond to human input but also will be able to cater to the emotional needs of the user. A chatbot is not supposed to respond "that's great to hear" to the user's input "I feel sick today". Up to our knowledge, there is no existing way that can automatically evaluate the performance of a chatbot in terms of empathetic responding. In this research, we aim to train an evaluator that can effectively evaluate the emotional consistency of chatbots.

### 1.2 Related Work

**Empathetic dialogues** There are studies (Rashkin et al., 2019; Li et al., 2017; Zhou et al., 2018; Sheen, 2021) that provide empathetic dialogues that we can train our evaluator upon.

**Chatbots** There are tens if not hundreds of chatbots available online. We choose two main-stream ones, namely, Blenderbot[3] by Facebook, and DialoGPT-large[4] by Microsoft, and evaluate their emotional consistency using our proposed evaluator.

### 1.3 Our Work

In this research, we propose an evaluator that can effectively evaluate chatbots' performance in terms of empathetic responding. First, we come up with a Neutral/Non-Neutral classifier (NNC) to determine whether the user's utterance is emotional or not. Second, we train an Emotion Classifier (EC) to determine whether an utterance is positive or negative emotionally and evaluate the emotional consistency of chatbots. Third, we validate the efficiency of EM by comparing the evaluations of EC with human evaluations on an evaluation data set. Finally, we use the validated EC to evaluate the emotional consistency of two main-stream chatbots (Blenderbot by Facebook, and DialoGPT-large by Microsoft) using a test data set, and compare their performances in terms of generating emotionally consistent conversations.

---

[1] https://github.com/OsirisLambert/chatbot-evaluator
[2] Youtube Demonstration:
https://www.youtube.com/watch?v=x66Alclp3uI
[3] https://huggingface.co/facebook/blenderbot-400M-distill
[4] https://huggingface.co/microsoft/DialoGPT-large

## 2 Methods

To evaluate the emotional consistency of a chatbot, we need to first have a Neutral/Non-Neutral classifier (NNC) to distinguish whether an utterance is with emotion or not. After we are left with those sentences that are non-neutral, we need another Emotion Classifier (EC) to classify whether a sentence is with positive emotion or negative emotion.

### 2.1 Datasets

The dataset we use to train the NNC is combined with empathetic data (Rashkin et al., 2019), DailyDialog (Li et al., 2017), MEISD (Zhou et al., 2018), and counsel data (Sheen, 2021). After combining those datasets into a single one, we split the training, validation, and testing set with ratios of 80%, 10%, and 10%. For EC, we use the same dataset and methods as we did in NNC, only that now we exclusively consider those utterances with non-neutral emotions.

In our implementation, we find that some labels are difficult to be regarded as positive or negative emotions, such as "surprise". To solve this problem, we count the number of positive utterances and negative utterances with each label by using an available online positive/negative classifier[5]. If the proportion of positive utterances with the label is greater than 70%, this label is classified as a positive emotion. If the proportion of negative utterances with the label is greater than 70%, this label is classified as a negative emotion. For labels that do not meet the above two conditions, we evaluate each utterance with these labels whether each utterance is with positive or negative emotion, not at the label level.

For the purpose of the final evaluation, we ask 5 people to chat with each other and generate a Human Conversation dataset (Appendix A) that consists of 141 dialogue turns. We use this dataset to ensure our evaluator's accuracy by comparing the performance between human-to-human dialogues and human-to-chatbot dialogues.

### 2.2 Classifiers

#### 2.2.1 Neutral/Non-Neutral Classifier[6]

We use the Hugging Face transformer (Wolf et al., 2020) to train this text-classification. We use a pre-trained model, 'roberta-base' (Conneau et al., 2020), as the baseline model. We fine tune hyperparameters to: max sequence length=128, training batch size per device=32, learning rate=1e-5, training epochs=3. We get 93.98% accuracy on the evaluation dataset and 91.92% accuracy on the test dataset.

---

[5]https://huggingface.co/distilbert-base-uncased-finetuned-sst-2-english
[6]https://huggingface.co/Osiris/neutral_non_neutral_classifier

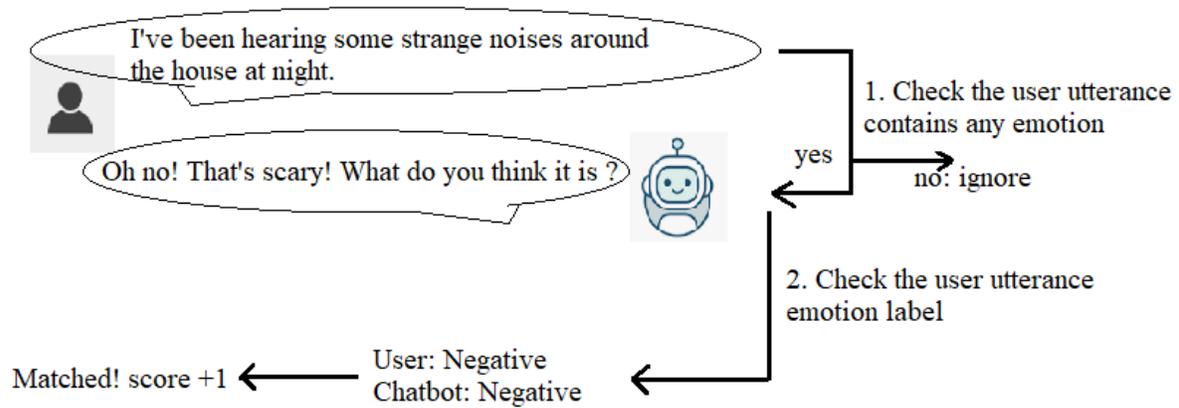

Figure 1. Flowchart for chatbots evaluator

### 2.2.2 Emotion Classifier[7]

Similar to the Neutral/Non-Neutral classifier, we use the Hugging Face transformer to train emotion classifiers. We use the same hyperparameters as mentioned in Section 2.2.1. We get 83.82% accuracy on the evaluation dataset and 84.42% accuracy on the test dataset.

## 3 Experiments

We try to restore the real conversation scene while evaluating chatbots. We use NNC to determine whether users' utterances consist of any emotion. Next, we pass all emotional utterances into chatbots to generate corresponding responses. We put both user utterances and chatbot responses into EC to get their emotion labels. We assign a score for each turn (one user utterance and one chatbot response) based on their emotion label (Figure 1).

---
[7] https://huggingface.co/Osiris/emotion_classifier

### 3.1 Chatbots

We totally adopted two chatbots to generate our system response, Blenderbot-400M-distill and Dialo-GPT-large. Blenderbot is a well-trained chatbot by Facebook and has a long-term memory. The chatbot model blends a diverse set of conversational skills, like empathy, knowledge, and personality together in one system. We use the 400M-distill version of Blenderbot which was built with 400M parameter neural models.

DialoGPT is a SOTA large-scale pretrained dialogue response generation model for multi-turn conversations by Microsoft. It extends GPT-2 to address the challenges of conversational neural response generation. Here, we use Large-sized(762M) GPT-2 models with 3 epochs, the largest version of DialoGPT.

## 3.2 Functions & Codes

We create three functions to process the raw dialogues, which are load_neutral / load_non_neutral, and load_emotion. We also have a chatbot class that can create our chatbots.

The basic logic of the two load_neutral / load_non_neutral functions is to apply NNC to each sentence from the original dataset. If it turns out to be neutral, then put it into the neutral sub dataset, and vice versa.

In the chatbot class, the required parameters are tokenizer and model, which determine the chatbot to be used. When the Blenderbot-400M-distill is chosen, the generate_response function should be used to generate a response given a user utterance; While when the DialoGPT-large is chosen, the generate_response_gpt function should be used. The reason why we have two generate_response functions is that the decoding of output tensors for the two chatbots are slightly different. The generate_response function will be applied to a dataset that only contains non-neutral user utterances and generates chatbot utterances. A new column is created that consists of those chatbot utterances.

In the load_emotion function, the main idea is to evaluate whether the user utterances and the chatbot responses are emotionally consistent for each turn in the dataset and give a score. The function returns a score which is the average of the score for each turn. The first step is to apply NNC to the chatbot response to see whether it is neutral or not. If it is neutral, we don't care whether the user's utterance is positive or negative, and we add 0.5 points directly. If the chatbot response is with emotion, then apply EC to both user utterance and chatbot response to see if they are emotionally consistent. If they are both positive or negative, add 1 point, otherwise, no extra points are added. Finally, go through all the turns in the dataset and return an average score.

We come up with an evaluator that integrates all the above functionalities and can directly yield evaluation scores for different chatbots.

## 3.3 Evaluation

We evaluate the performance of our evaluator by comparing its evaluation results with human evaluations, and we use a rating mechanism with a binary score distribution of 0 and 1 (0 means emotionally-inconsistent, and 1 means consistent). After rating all the dialogue turns in the evaluation dataset, we take the average score. If the average score by our evaluator matches well with human evaluations, it can prove that our evaluator is able to effectively evaluate the performance of a chatbot in terms of empathetic responding.

## 4 Results & Discussion

### 4.1 Evaluation Performance

We use the Human Conversation Dataset to test the performance of different chatbots, and we compare overall scores with human responses scores. (Table 1)

| Response by | Evaluator Score (sentence level) | Human Rating (sentence level) | Human Rating (dialogue level) |
|---|---|---|---|
| Human | 0.8085 | 0.8511 | 0.8942 |
| Blenderbot | 0.8865 | 0.8617 | 0.9117 |
| DialoGPT-large | 0.6879 | 0.6525 | 0.6442 |

Table 1: Human Rating v.s. Evaluator

We calculate the correlation score between evaluator and human score and get a result of 0.9375. It indicates that our evaluator is highly correlated to our human rating, which makes it a well-performed chatbot evaluator.

We also evaluate chatbots at the conversation level. The evaluator and human ratings are shown in detail in appendix B. We use the mean of minimum and maximum aggregation over a conversation and get a correlation score of 0.8882. Even though this score is lower than that of turn level, it still strongly indicates that our evaluator score is highly related to human rating.

These 2 scores together illustrate that our evaluator does a good job on evaluating the emotional consistency of a chatbot.

### 4.2 Test Dataset Performance

We use the testing set (section 2.1) to demonstrate the evaluations for 2 chatbots (Table 2).

| Response by | Evaluator Score |
|---|---|
| Blenderbot | 0.7926 |
| DialoGPT-large | 0.7063 |

Table 2: Evaluator score for chatbots

We find that the evaluator score for Blenderbot is better than the score for DialoGPT-large. This result is reasonable because Blenderbot can generate more emotionally consistent responses than DialoGPT-large, by human evaluation.

### 5 Conclusions

We combine four datasets available online. We train two classifiers, one is the Neutral/Non-neutral Classifier with an accuracy of 91.92% on the test dataset, and the other is the Emotion Classifier with an accuracy of 84.42% on the test dataset. We also create an evaluator to evaluate whether the chatbots' responses are emotionally consistent or not and give a score. Finally, we evaluate our evaluator by comparing human evaluations and our evaluator evaluations. We hope that our results will inspire more people working on this evaluator field, and help researchers to generate more emotionally consistent chatbots.

We can continue to improve our Emotion Classifier. Currently, our EC is a binary classifier that can only classify whether an utterance is positive or negative. In the future, we can use a multi-label dataset to train the Emotion Classifier, which can have a more accurate judgment on the emotion of each utterance.

In dialogue level evaluation, we encourage people to improve our approach by getting a more precise scoring method.

In addition, we can add more evaluation datasets. For those datasets, we can purposely negate the emotion to make it as bad as possible. By adding those negative samples, we can check the evaluator performance in the opposite way.

# Appendix

## A Human Conversation Dataset Details

In the human conversation dataset, we count the average turns for a conversation is about 2 turns. Table 3 shows frequency of different turns in the dataset.

Table 4 provides average utterance length for all types of utterances.

| # of Conversation Turns | Frequency |
|---|---|
| 1 | 2 |
| 2 | 43 |
| 3 | 7 |
| 4 | 8 |

Table 3: Conversation Turns

| Type of Utterance | Ave. Utterance Length |
|---|---|
| User Utterances | 12.32 |
| Human Response | 10.72 |
| Blenderbot Response | 15.74 |
| DialoGPT Response | 7.65 |

Table 4: Utterances Length

## B Human Conversation Dataset Dialogue Level Analysis

Although our evaluator gives a score based on sentence level utterance, we also provide good statistics on dialogue level utterances. Table 5 gives our experiment results. The techniques we used to calculate scores are maximum aggregation, minimum aggregation, and middle score between those two techniques.

| Response by | Evaluator Score (dialogue level-max) | Evaluator Score (dialogue level-min) | Evaluator Score (dialogue level-mid) |
|---|---|---|---|
| Human | 0.9492 | 0.5847 | 0.7669 |
| Blenderbot | 0.9831 | 0.7627 | 0.8729 |
| DialoGPT-large | 0.9068 | 0.4237 | 0.6653 |

Table 5: Dialogue Level Evaluator Scores